\documentclass[10pt,twocolumn,letterpaper]{article}

\usepackage{cvpr}
\usepackage{times}
\usepackage{epsfig}
\usepackage{graphicx}
\usepackage{amsmath}
\usepackage{amssymb}
\usepackage{algorithm}
\usepackage{algorithmic}
\usepackage{bm}
\usepackage{color}
\usepackage{url}
\usepackage{multirow}
\usepackage[table,xcdraw]{xcolor}

\usepackage[pagebackref=true,breaklinks=true,letterpaper=true,colorlinks,bookmarks=false]{hyperref}

\cvprfinalcopy 

\def\cvprPaperID{5549} 

\ifcvprfinal\fi
\begin{document}

\title{Adaptive Weighting Multi-Field-of-View CNN \\ for Semantic Segmentation in Pathology}

\author{Hiroki Tokunaga$^\text{1}$~~~~~Yuki Teramoto$^\text{2}$~~~~~Akihiko Yoshizawa$^\text{2}$~~~~~Ryoma Bise$^\text{1,3}$\\
$^\text{1}$Kyushu University, Fukuoka, Japan~~~ 
$^\text{2}$Kyoto University Hospital, Kyoto, Japan~~~\\ 
$^\text{3}$Research Center for Medical Bigdata, National Institute of Informatics, Tokyo, Japan\\ 
}


\maketitle

\begin{abstract}
Automated digital histopathology image segmentation is an important task to help pathologists diagnose tumors and cancer subtypes.
For pathological diagnosis of cancer subtypes, pathologists usually change the magnification of whole-slide images (WSI) viewers. A key assumption is that the importance of the magnifications depends on the characteristics of the input image, such as cancer subtypes.
In this paper, we propose a novel semantic segmentation method, called Adaptive-Weighting-Multi-Field-of-View-CNN (AWMF-CNN), that can adaptively use image features from images with different magnifications to segment multiple cancer subtype regions in the input image.
The proposed method aggregates several expert CNNs for images of different magnifications by adaptively changing the weight of each expert depending on the input image. 
It leverages information in the images with different magnifications that might be useful for identifying the subtypes.
It outperformed other state-of-the-art methods in experiments.
\end{abstract}

\vspace{-4mm}
\section{Introduction}
Automated digital pathology image analysis is an important task to help pathologists diagnose tumors and cancer subtypes.
In particular, many segmentation methods, such as for segmenting tumor regions, have been proposed. The state-of-the-art methods accurately distinguish regions (normal and tumor) in digital pathology images~\cite{ChenH2016}\cite{XuY2017}\cite{XuY2016}.
Recent progress in medicine has emphasized, the importance of cancer subtype analysis in histology. For example, Yoshizawa {\it et al.}~\cite{YoshizawaA2011} showed that knowledge of lung adenocarcinoma subtypes could be used to predict the prognosis of a patient who underwent surgical resection statistically. 
The development of multiple-subtype segmentation will be important for pathological image analysis.

\begin{figure}[t]
\includegraphics[width=\linewidth]{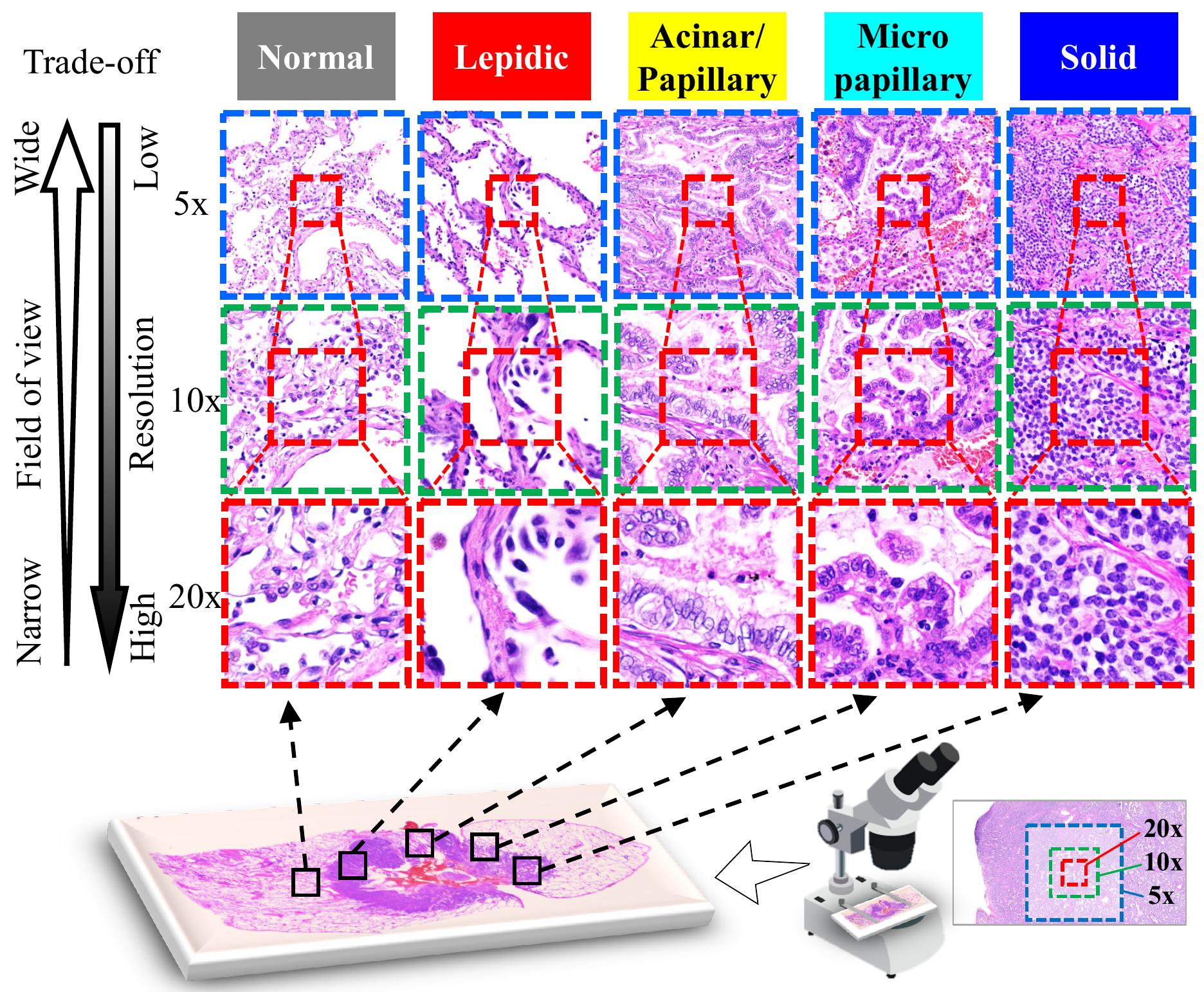}
\caption{Examples of pathology images at difference magnifications; Top: 5x, Middle: 10x, Bottom: 20x. The columns indicate the subtypes: from left to right, Normal, Lepidic, Acinar/Papillary, Micropapillary, and Solid.}
\label{fig:TradeOff}
\vspace{-3mm}
\end{figure}

Convolutional neural networks (CNN)~\cite{KrizhenvskyA2012} have been used for classification and segmentation tasks and they have been shown to outperform traditional computer vision techniques in various applications.
Whole-slide images (WSI), which are often used in digital pathology, cannot be inputted to a CNN because they are so large (e.g., 100,000 $\times$ 50,000 pixels) compared with a natural image (i.e., WSI is over 10000 times the size of a natural image).
Therefore, most methods take a patch-based classification approach that first segments a large image into small patches and then classifies each patch~\cite{WangD2016}. 
However, a small patch image has less context of a wide range of texture patterns that might be useful for classification.
In order to extend the field of view under the size limit, the input image is usually downsampled. As a result, the spatial resolution is reduced (Figure~\ref{fig:TradeOff}).
This trade-off between the size of the field of view and the resolution of the input image makes it difficult to segment cancer subtypes from an input patch image.

On the other hand, to make pathological diagnoses, pathologists usually check images by changing their magnification in WSI viewers (i.e., they use several different scaled images). 
They can check a wide range of texture patterns in a low-magnification image, whereas they use high-magnification images to check details, such as the shapes of individual cells that are too small to be clearly seen in the low-resolution images (Figure~\ref{fig:TradeOff}).
It means that it is important for diagnosis to use both of high-resolution images with the narrow field of view and low-resolution images with the wide field of view.
A key assumption is that the importance of the magnifications depends on the characteristic of the input image.
For example in Figure~\ref{fig:TradeOff}, images with a wide field of view (top) have more discriminative features than the narrow field of view with high-resolution (bottom) for distinguishing acinar/papillary and lepidic subtypes.
On the other hand, high-resolution images (bottom) have more discriminative features than the wide field of view (top) for distinguishing normal and lepidic subtypes. 
This indicates that it is important to adaptively use the images with different magnification depending on the input image.

In summary, the main contributions of our study are:
\vspace{-2mm}
\begin{itemize}
\item We propose a semantic segmentation method that can aggregate contextual information from multiple magnification images by adaptively weighting several segmentation networks (expert CNNs) that are trained using different-magnification images. Our method leverages the contexts from both wide field-of-view and high-resolution images that might be useful for identifying the subtypes.
\vspace{-2mm}
\item Our end-to-end learning re-trains the expert CNNs so that all experts work complementarily to increase the cases that either expert can predict a correct answer, and trains an aggregating CNN to be able to adaptively aggregate the predicted results. This contributes to improving the segmentation performance.
We also analyzed the effectiveness of the learning by comparing the prediction results of experts between the before and after the end-to-end learning.
\vspace{-2mm}
\item We demonstrate the effectiveness of our method on a challenging task; segmentation of subtype regions of lung adenocarcinomas. Our method outperformed the state-of-art methods, in particular, in a multi-class segmentation task. We also show that our method can be applied to any type of network.
\end{itemize}

\section{Related works}
Many methods have been proposed for pattern recognition tasks in pathology: Patient-level and WSI-level pN-stage estimation (CAMELYON 2016 and 2017)~\cite{Camelyon2017}\cite{Camelyon2016}, and segmentation of tumors.
Segmentation methods distinguish the tumor regions from normal regions in a pathological image (WSI), and they can be roughly classified into patch-wise classifications and pixel-wise semantic segmentations.

\noindent {\bf Patch-based approach:} The patch-based methods segment the large WSI into small image patches and then classify each patch image~\cite{AltunbayD2010}\cite{ChangH2014}\cite{CruzRoaA2014}\cite{MousaviHS2015}\cite{XuY2015}\cite{ZhouY2014}. 
Wang {\it et al.}~\cite{WangD2016} used a CNN to extract features from each patch and assign a prediction score. They performed breast metastasis cancer detection based on the predicted score map in WSIs of sentinel lymph node biopsies. 
Hou {\it et al.}\cite{HouL2016} proposed an EM-based classification method with a CNN that automatically identifies discriminative patches by utilizing the spatial relationships of patches. It was used to classify glioma and non-small-cell lung carcinoma cases into subtypes.
As discussed in the introduction, contextual information from image features of a fixed patch size is not enough to identify the cancer type. To address this shortcoming, several methods that incorporate multi-scale contextual information into a patch-wise classification have been proposed~\cite{AlsubaieN2018}\cite{KongB2017}\cite{SirinukunwattanaK2018}.
These methods are efficient for rough segmentation (i.e., patch-level segmentation). To obtain the pixel-level segmentation, the sliding window strategy is required. 
One of the drawbacks is its slowness; the classification process must be run separately on each pixel, and there is a lot of redundancy due to overlapping patches.

\noindent {\bf Semantic Segmentation:} To overcome the above shortcoming, FCN produces a segmentation mask image with high resolution; its architecture consists of downsampling layers for extracting image features and upsampling layers for obtaining a segmentation mask~\cite{LongJ2015}.
U-net~\cite{RonnebegerO2015} is widely used for segmentation problems; it introduces skip connections from downsampling layers to upsampling layers to preserve the information in high-resolution images. 
This network won the ISBI challenge 2015 for segmentation of neuronal structures in electron microscopy. 
Many semantic segmentation methods have been proposed for natural image analysis~\cite{GouldS2009}\cite{KohliP2009}\cite{YaoJ2012}.
In particular, graphical model~\cite{ChenLC2017_TPAMI}\cite{JampaniV2016}\cite{ZhengS2015}\cite{VemulapalliR2016}, spatial pyramid pooling~\cite{ChenLC2017_TPAMI}\cite{ChenLC2017_arXiv}\cite{ZhaoH2017}, dilated convolution~\cite{YuF2016} and multi-scale inputs (i.e., image pyramid)~\cite{ChenLC2017_TPAMI}\cite{ChenLC2016_cvpr}\cite{EigenD2015}\cite{LinG2016}\cite{PinheiroP2014} exploit contextual information for segmentation. These models have shown promising results on several segmentation benchmarks by aggregating multi-scale information. 
They assume that an entire image can be inputted to a single network; the entire image is scaled to change the range of feature extraction.
In this scaling, where the field-of-view of the scaled images are same.
However, as we discussed in the introduction, the trade-off between spatial resolution and the size of the field of view remains a problem for semantic segmentation in pathology because WSIs are huge and the input to a single network is limited due to the size of the GPU memory size.
In this study, we incorporate multi-field-of-view and multi-resolution contextual information into a pixel-wise semantic segmentation scheme.

\noindent {\bf Weighting (Gating):} We propose an aggregation method that can adaptively weight multiple CNNs trained with different field-of-view images.
There are several methods that can adaptively weight the effective channels~\cite{HuJ2018}, pixels (location)~\cite{LiM2018}, and scales~\cite{DingH2018}\cite{QinY2018} in a single network.
Hu {\it et al.}~\cite{HuJ2018} proposed SENet which adaptively weights the channel-wise feature responses by explicitly modeling interdependencies between channels. Their network improves state-of-the-art deep learning architectures.
Sam {\it et al.}~\cite{SamDB2017} proposed a hard-switch-CNN network that chooses a single optimal regressor from among several independent regressors. It was used for counting the number of people in patch images. Kumagai {\it et al.}~\cite{KumagaiS2017} proposed a mixture of counting CNNs for the regression problem of estimating the number of people in an image.

The studies most related to ours are follows.
Alsubaie {\it et al.}~\cite{AlsubaieN2018} proposed a multi-resolution method that simply inputs multi-field-of-view images as multi-channels into a single network. The method has slightly higher accuracy compared with that of a single scale network.
Sirinukunwattana {\it et al.}~\cite{SirinukunwattanaK2018} systematically compared different architectures. They trained each CNN on images with different magnifications, and then fused the results from the CNNs in several ways, such as CNN, LSTM, to obtain the final results. These methods are for classifying patch images; they cannot be directly applied to semantic segmentation CNNs. In addition, they do not take into account that the importance of the magnifications depends on an input image.

Unlike these current methods, our novel neural network architecture and learning algorithm can adaptively use the features from the multi-field-of-view images depending on the characteristic of the input image for semantic segmentation in pathology.

\section{Effect of field of view}
In this section, we will investigate how the contextual information (resolution and field-of-view) is related to the discriminative features for segmenting cancer subtypes.
We compared the outputs from CNNs that were trained individually using images with different magnifications (20x, 10x, 5x) that have a trade-off between their resolution and the field-of-view size.

In the experiment, we used lung adenocarcinoma images annotated by pathologists. 
We trained expert networks 1, 2 and 3 by using images with the different magnifications as shown in the top, middle and bottom row in Figure~\ref{fig:TradeOff} respectively.
The details of the dataset and setup are described in Section~\ref{Experiments}.

\begin{figure}[t]
\centering
\includegraphics[width=0.9\linewidth]{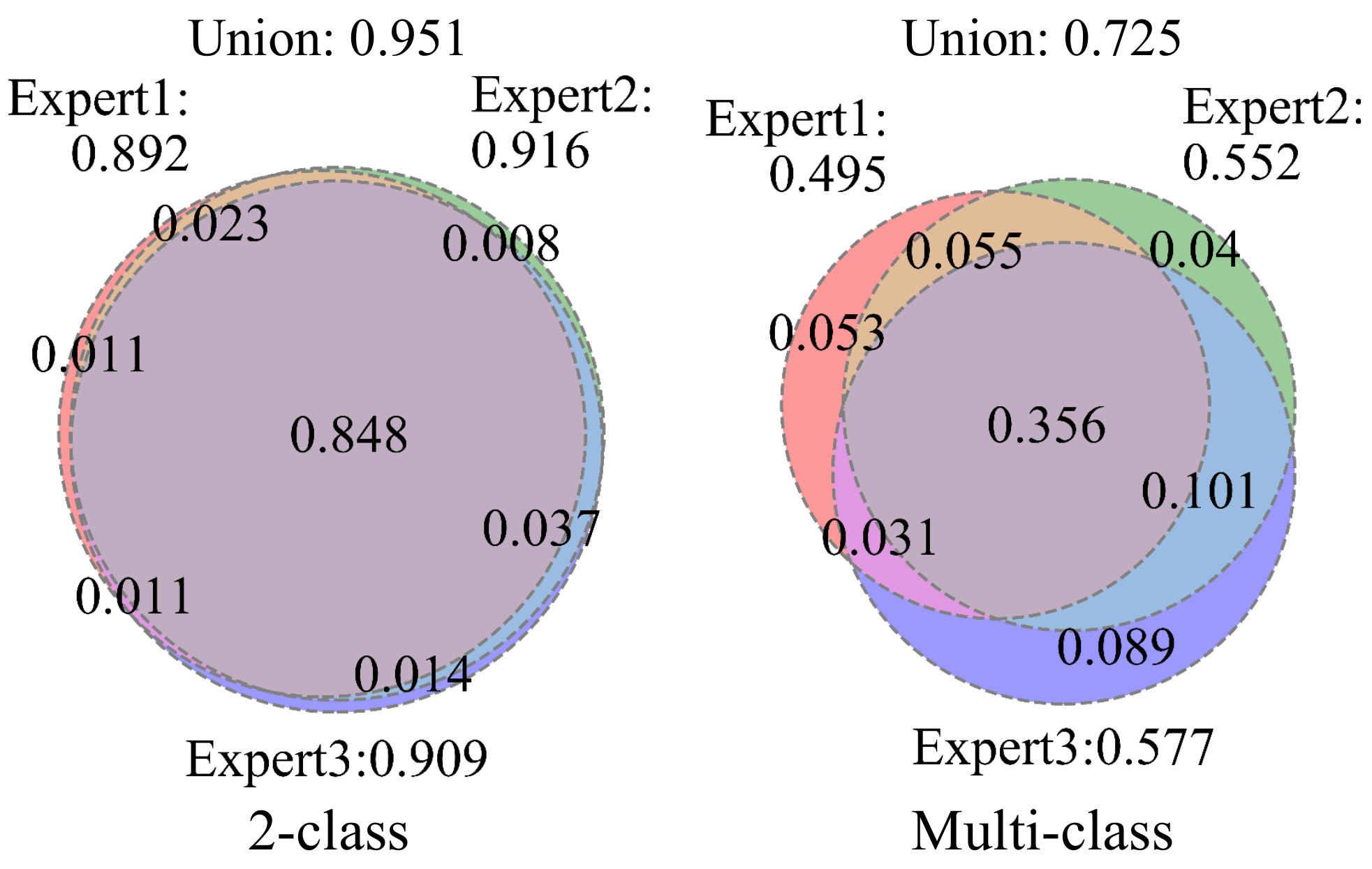}
\caption{Venn diagrams of correct answer rates for individually pre-trained expert CNNs on two-class (left) and multi-class (right) segmentation tasks.}
\label{fig:Venn_3experts_before}
\vspace{-4mm}
\end{figure}

In Figure~\ref{fig:Venn_3experts_before}, a circle indicates a set of pixels that an expert correctly predicted, and the value indicates the ratio of pixels correctly predicted to all pixels.
The overlap region of the three experts indicates the set of pixels that all experts correctly predicted.
The non-overlap region indicates the pixels that only one expert correctly predicted. Union indicates the set of pixels that either expert predicted a correct class.
This Venn diagram shows the similarity of the predicted results of the expert networks. 

The left figure showing the results of the two-class segmentation task indicates that the experts gave very similar results.
The right figure showing the results of the multi-class segmentation task indicates that the non-overlap areas were large, and thus, the union of the experts was much larger than a single expert (by over 15$\%$). 
This indicates that the contextual information from different-magnification images is effective for multiple subtype semantic segmentation (i.e., our assumption that the importance of the magnifications depends on an input image is reasonable.).

These results indicate that if a method can adaptively aggregate experts depending on the characteristic of the input image, the method can outperform the single experts, in which the accuracy is expected to close to the accuracy of the union of the experts.
In addition, if a method can re-train the experts so that each expert is specialized to make the difference large and the size of the union increases, the method will be able to outperform than the union of the original individual experts.
To achieve this goal, we select the mixture of experts approach that aggregates the expert CNNs for different-magnification images while adaptively changing the weight of each CNN depending on the input image. The details of this method are presented in the next section.

\begin{figure}[t]
\label{fig:Adaptive Weighting Multi-FOV CNN}
\begin{center}
\includegraphics[width=0.9\linewidth]{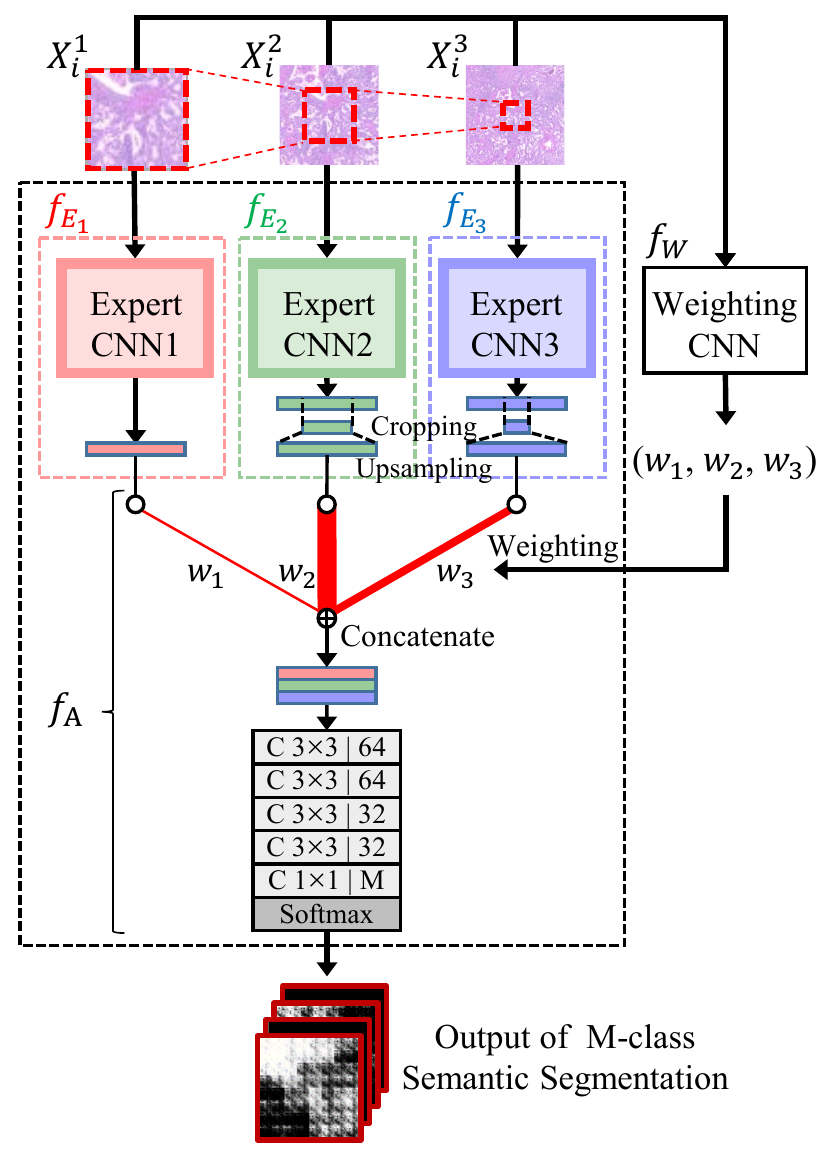}
\caption{Overview of Adaptive-Weighting-Multi-Field-of-View-CNN architecture. Red dotted boxes on the input images are the target regions of semantic segmentation. After each expert CNN makes a prediction, the cropped target area is upsampled to the same size of $X_i^1$.}
\end{center}
\vspace{-4mm}
\end{figure}

\section{Proposed method}
\label{sec:Adaptive Weighting Multi-FOV CNN}
Figure~\ref{fig:Adaptive Weighting Multi-FOV CNN} shows an overview of our Adaptive-Weighting-Multi-Field-of-View-CNN (AWMF-CNN).
To address the trade-off between the resolution and the size of the field of view, we use three different-magnification images $\bm{X}_i$ ($X_i^1$ for 20x, $X_i^2$ for 10x, $X_i^3$ for 5x) as inputs, where each input image has different spatial resolutions and fields of view. $i$ indicates an index of an image data set. Here, $X_i^1$ is the target image patch that will be segmented, which has the highest spatial resolution with the narrowest field of view.
$X_i^2$ and $X_i^3$ are low magnification images of $X_i^1$, where the image center area (red dotted boxes) are the same as shown in Figure~\ref{fig:Adaptive Weighting Multi-FOV CNN}.
In addition to the target area, the image features from the peripheral regions, which is the outer regions of the red box in $X_i^2$, $X_i^3$, are used to segment the multiple-subtype cancer regions in $X_i^1$.
Our network can adaptively estimate the weights of three different magnifications depending on an input image and aggregate these image features by using the estimated weights to segment cancer-subtypes.

\subsection{Network architecture}
\label{sec:NetworkArchitecture}
Our AWMF-CNN consists of three types of network: expert CNNs ($f_{E_1}$, $f_{E_2}$, $f_{E_3}$), a weighting CNN $f_{W}$, and an aggregating CNN $f_{A}$ as shown in Figure~\ref{fig:Adaptive Weighting Multi-FOV CNN}.
We use the U-net architecture~\cite{U-net}\cite{RonnebegerO2015} for each expert CNN.
The CNNs are trained such that each one becomes specialized for segmenting images of a particular magnification in pre-training.
Each channel image in the output layer of each network is a heat map of the likelihood of each subtype in the target region, where the number of the output channels equals the number of subtypes to be segmented.
Since the field of views in $X_2^i$ and $X_3^i$ are different from that of the target region $X_1^i$, the target region (red box) of each output heat map is cropped and upsampled to the same size as the output of $f_{E_1}$.
Here, the cropped heat map is estimated also using the peripheral context i.e., image features from the outer regions of the target area (outside of the red box).
The outputs of these experts are aggregated on the aggregating CNN to produce the final results.

Under the assumption that the importance of magnification images differ depending on the input image, we developed a weighting CNN that adaptively estimates the weights of the expert CNNs by using the input images.
We modified Xception~\cite{CholletF2016} (developed by Google) for classification by replacing the fully connected (FC) layers with global average pooling, a 3-class FC layer, and a sigmoid activation function to output three weights and fine-tuned the network by using the trained parameters for 1000-class classification as the initial values. The range of each weight is $0$ to $1$.

The aggregating CNN concatenates the outputs of expert CNNs with the estimated weights. This network outputs the final segmentation result of the target region.
It has a simple architecture consisting of five convolutional layers and a softmax function as shown in Figure~\ref{fig:Adaptive Weighting Multi-FOV CNN}. Each convolutional layer except for the last is followed by a batch normalization~\cite{ioffe2015batch} and an exponential linear unit function~\cite{clevert2015fast}.

\subsection{Training algorithm}
\label{sec:Training}
Algorithm~\ref{alg1} is an overview of the training procedure.
We use two training data sets for the weighting CNN $f_{W}$ and other CNNs ($f_{E_1}$, $f_{E_2}$, $f_{E_3}$, $f_{A}$).
$N_X$ training image patches $\{X_i^1, X_i^2, X_i^3\}_{i=1}^{N_X}$ with a ground-truth segmentation map $\{T_{i,c}^{1}, T_{i,c}^{2}, T_{i,c}^{3}|c=1,...,M\}_{i=1}^{N_X}$ are used to train ($\{f_{E_k}\}_{k=1}^3$, $f_A$), and $N_{X'}$ training images $\{{X'}_i^1, {X'}_i^2, {X'}_i^3\}_{i=1}^{N_{X'}}$ with $\{{T'}_{i,c}^{1}, {T'}_{i,c}^{2}, {T'}_{i,c}^{3}|c=1,...,M\}_{i=1}^{N_{X'}}$ are used to train the weighting CNN $f_{W}$ where $M$ is the number of classes.
The image patches $\{X_i^1, X_i^2, X_i^3\}$ are different-magnification images that contain the same target regions, as explained above.

\noindent {\bf (0) Initialization (Pre-train)}

The three expert CNNs $\{f_{E_k}\}_{k=1}^3$ are pre-trained independently to estimate the heat maps for each magnification using different training sets.
For the loss function of the experts, we used the sum of weighted cross-entropy terms for each spatial position (pixel) in the CNN output map. The loss functions are defined as follows:
\begin{eqnarray}
\hspace{-8mm}&&Loss_{E_k} \!=\! - \sum_{i=1}^{N_X} \sum_{j \in X_i^k} \sum _{ c = 1 } ^ { M } \alpha _ { c } T_{i,c}^{k}(j) \log Y_{i,c}^k(j), \\
\hspace{-8mm}&&\alpha_c \!=\! \frac{Number~of~all~pixels}{M~\times~Number~of~pixels~of~class~c},
\label{eq:LossE}
\end{eqnarray}
where $c$ is the class index, $X_i^k$ is the input image patch for the $k$-th expert, $j$ is the $j$-th pixel of $X_i^k$, $T_{i,c}^{k}(j)$ is th ground-truth of the $j$-th pixel of $T_{i,c}^{k}$ from manually labelled annotations, $Y_{i,c}^k(j)$ is the prediction of the network, and $\alpha_c$ is the weight for eliminating bias due to the imbalance in the number of images in different classes. $Y_{i,c}^k$ for $k=2,3$ is the output of the CNN before cropping. 
The loss is optimized by back-propagating the CNN via the optimizer for each network and the network parameters $\{ \Theta_k\}_{k=1}^3$ are updated.
In the initialization training, each expert CNN is specialized for images of a specific magnification.

\begin{algorithm}[t]
\caption{AWMF-CNN training algorithm}
\label{alg1}
\begin{algorithmic}[1]
\STATE Input: $N_X$ training image patches $\{X_i^1, X_i^2, X_i^3\}_{i=1}^{N_X}$ with ground truth $\{T_{i,c}^{k}\}_{i,c,k}$, and $N_{X'}$ training images $\{{X'}_i^1, {X'}_i^2, {X'}_i^3\}_{i=1}^{N_{X'}}$ with $\{{T'}_{i,c}^{k}\}_{i,c,k}$
\STATE \% Initialization: Pre-training for $f_{E_1}, f_{E_2}, f_{E_3}$
\STATE Backpropagating to train $\{ \Theta_k^{(0)}\}_{k=1}^3$ using $\{ T_{i,c}^k\}_{k=1}^3$ respectively
\STATE \% Training for $L$ epochs
\FOR{$l=1$ to $L$}
\STATE \% Generate training data for $f_{W}$
\FOR{$i=1$ to $N_{X'}$}
\STATE \% Output $f_{E_k}$ with input ${X'}_i$
\STATE ${Y'}_{i,c}^k$ = $f_{E_k}({X'}_i^k; \Theta_k^{(l-1)})$
\STATE $w_i^k = \frac { 2 | {Y'}_{i,c}^k \cap {T'}_{i,c}^{k} | } { | {Y'}_{i,c}^k | + | {T'}_{i,c}^{k} | }$, \hspace{2mm}$\bm{w}_i^{(l)} = [w_i^1, w_i^2, w_i^3]$
\ENDFOR
\STATE $S_{train} = \{ \bm{{X'}}_i, \bm{w}_i \}_{i=1}^{N_{X'}}$
\STATE \% Training $f_{W}$ for 1 epoch
\STATE Train $f_{W}$ with $S_{train}$ and update $\Theta_{W}^{(l)}$
\STATE \% Train ($\{f_{E_k}\}_{k=1}^3$, $f_A$) with $f_{W}$
\FOR{$i=1$ to $N_X$}
\STATE \% Estimate weights $\bm{w}_i$ by $f_{W}$ with current $\Theta_{W}^{(l)}$
\STATE $\bm{w}_i^{(l)} = f_{W}(\bm{X}_i; \Theta_{W}^{(l)})$
\STATE \% Train $\{\Theta_k\}_{k=1}^3$ and $\Theta_A$ with $\bm{w}_i^{(l)}$
\STATE Backpropagating to train ($\{f_{E_k}\}_{k=1}^3$, $f_A$) with $\bm{w}_i^{(l)}$ and update $\{\Theta_k^{(l)}\}_{k=1}^3$ and $\Theta_A^{(l)}$
\ENDFOR
\ENDFOR
\STATE Output: trained parameters $\{ \Theta_k\}_{k=1}^3$ for $f_{E_k}$, $\Theta_{A}$ for $f_A$ and $\Theta_{W}$ for $f_{W}$
\end{algorithmic}
\label{alg:outerRoop}
\end{algorithm}

Two types of networks: the integrated network consisting of four networks ($\{f_{E_k}\}_{k=1}^3$, $f_A$) (black dotted box in Figure~\ref{fig:Adaptive Weighting Multi-FOV CNN}), and weighting CNN $f_{W}$, are alternately optimized by iteratively processing the following step. The initialized parameters are used in the first iteration.

\noindent {\bf (1-1) Generate Training Data for Weighting CNN}

First, to train the weighting CNN $f_{W}$, the training data for weight set is generated by using the training data $\{\{{X'}_i^k\}_{k=1}^3, \{{T'}_{i,c}^{k}\}_{c,k}\}_{i=1}^{N_{X'}}$.
The key idea in training the weighting CNN is that when a magnification of a test image has more discriminative features than the other magnification, the corresponding expert should produce a good estimate for the test image.
To estimate the weights, we use the {\it Dice coefficient} between the estimation image ${Y'}_{i,c}^k$ and the ground-truth ${T'}_{i,c}^{k}$ for each class $c$ as the weights of the experts,
\begin{equation}
 w_i^k = \frac { 2 | {Y'}_{i,c}^k \cap {T'}_{i,c}^{k} | } { | {Y'}_{i,c}^k | + | {T'}_{i,c}^{k} | }, 
 \hspace{5mm}\bm{w}_i^{(l)} = [w_i^1, w_i^2, w_i^3],
\end{equation}
where $\cap$ means the element-wise product and $|\cdot|$ means the sum of elements. $l$ is an iteration index.

\noindent {\bf (1-2) Train Weighting CNN}

The weighting CNN $f_{W}$ is trained using a set of training images and weights $S_{train}=\{\bm{X'}_i, \bm{w}_i\}_{i=1}^{N_{X'}}$, and the network parameters $\Theta_{W}$ are updated by backpropagating the CNN via optimizer: the loss function is the mean squared error (MSE) defined as follows:
\begin{equation}
Loss_{W} = \sum _ { i = 1 } ^ { N_{X'} } \sum _ { k = 1 } ^ { 3 } \left( w_i^{ k } - y_i^{ k } \right) ^ { 2 },
\label{eq:LossW}
\end{equation}
where $k$ is the index of the expert CNNs, and $y_i^k$ is the weight predicted by the weighting CNN.

\noindent {\bf (2) End-to-End Learning of the integrated network}

The integrated network consisting of $\{f_{E_k}\}_{k=1}^3$, $f_A$ are trained with weights estimated from the weighting CNN with the training data $\{\{X_i^k\}_{k=1}^3,\{T_{i,c}^{k}\}_{k=1}^3\}_{i=1}^{N_X}$ in end-to-end learning. 
The weights of each expert $\bm{w}_i^{(l)}$ are first estimated for a training image $\bm{X}_i$ by weighting CNN $f_{W}$ with the current $\Theta_{W}^{(l)}$. Using the estimated weights, the integrated network is trained in an end-to-end manner by backpropagating the CNN via the optimizer using the weighted cross-entropy loss,
\begin{equation}
Loss_{A}\!=\! - \sum_{i=1}^{N_X} \sum_{j \in X_i^k} \sum _{ c = 1 } ^ { M } \alpha _ { c } T_{i,c}(j) \log Y_{i,c}(j),
\end{equation}
\vspace{-5.5mm}
\begin{equation}
Loss = Loss_A + \sum_{k=1}^3{Loss_{E_k}},
\label{eq:Loss}
\end{equation}
where $Y_{i,c}(j)$ is the estimated score of the $j$-th pixel in the output image $Y_{i,c}$ of the aggregating layer. This training process is iterated for every training data $\bm{X}_i$ and $\{\Theta_k^{(l)}\}_{k=1}^3$, and $\Theta_A^{(l)}$ are updated.

This training algorithm is run until the maximum epoch or convergence.
Through it, each expert CNN becomes specialized for images in which the magnification is useful for segmentation.
The weighting CNN is trained to estimate the weights of the specialized experts depending on an input image.
The aggregating CNN is trained to estimate the final segmentation results of the target area, which aggregates the experts using the estimated weights.
Given a test image, the trained weighting CNN first estimates the weights, and then the trained integrated network predicts the final segmentation result.

\begin{figure}[t]
\centering
\includegraphics[width=\linewidth]{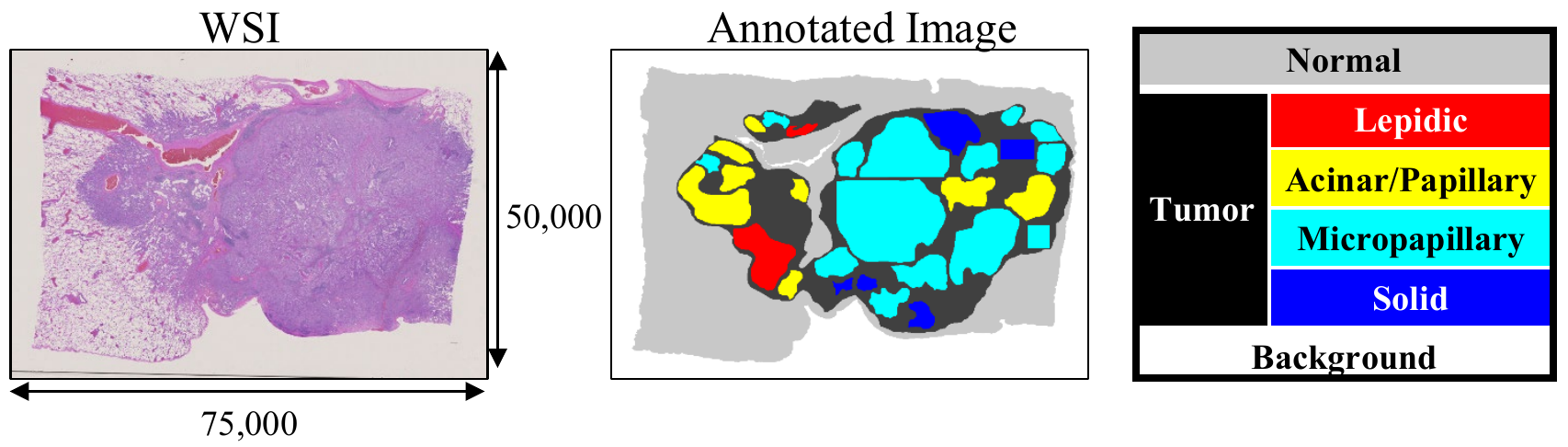}
\caption{Example of WSI. Left: original WSI ($50,000 \times 75,000$);
Middle: annotated image provided by pathologists; Right: annotation label and corresponding colors.}
\label{fig:Data}
\vspace{-3mm}
\end{figure}

\section{Experiments\label{Experiments}}
\begin{figure*}[t]
\centering
\includegraphics[width=0.96\linewidth]{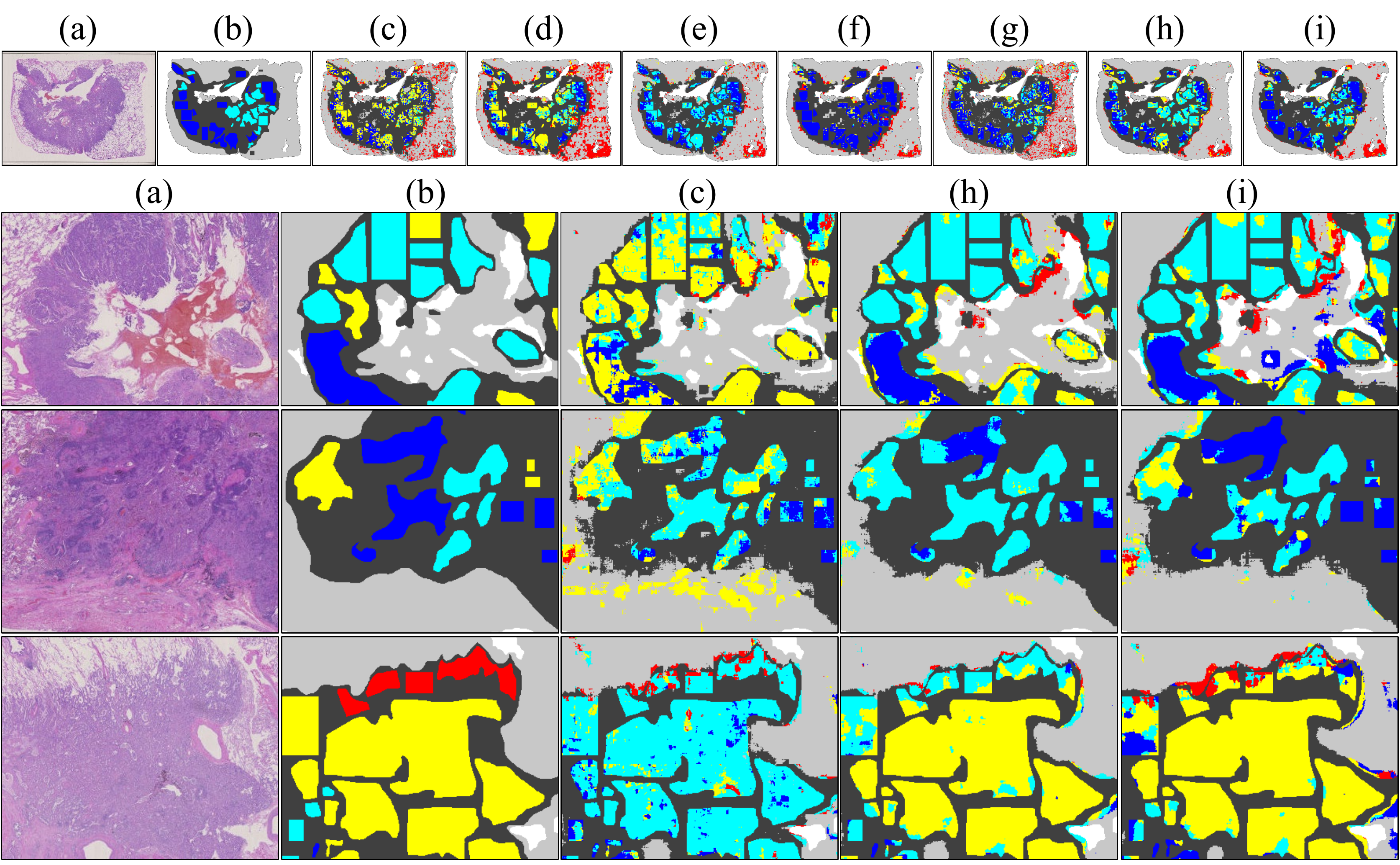}
\caption{Examples of segmentation results. 
(a) original images, segmentation images from (b) manual annotation, (c) U-net~\cite{U-net}\cite{RonnebegerO2015}, (d) SegNet~\cite{BadrinarayananV2017}\cite{Segnet-implement}, (e) Dilated-net~\cite{YuF2016}, (f) DeepLabv3+~\cite{ChenLC2017_arXiv}\cite{DeepLabv3plus}, (g) Hard-Switch-CNN~\cite{SamDB2017}, (h) Ours (Fixed), and (i) Ours (Adaptive). 
The color of the region indicates the subtype class (see Figure~\ref{fig:Data}). }
\label{fig:Results}
\vspace{-4mm}
\end{figure*}

We evaluated our method on two segmentation problems from whole slide images (WSI), including two-class segmentation into tumor and normal, and multi-subtype segmentation in lung adenocarcinoma.
For these experiments, we compared the segmentation accuracy with the following state-of-the-art methods: U-net~\cite{U-net}\cite{RonnebegerO2015},
SegNet~\cite{BadrinarayananV2017}\cite{Segnet-implement}, Dilated-net~\cite{YuF2016}, DeepLabv3+~\cite{ChenLC2017_arXiv}\cite{DeepLabv3plus}, and Hard-Switch-CNN (HS)~\cite{SamDB2017} that adaptively selects an expert network (does not aggregate multiple images).
For the proposed method, we evaluated two versions:
Ours (Adaptive) is Adaptive-Weighting-Multi-Field-of-View CNN (AWMF-CNN), and Ours (Fixed) is the Multi-Field-of-View CNN that uses the fixed weight 1.0 for aggregation, where the other setup was same as AWMF-CNN. 

\subsection{Dataset\label{dataset}}
Images of sliced lung adenocarcinoma stained by hematoxylin and eosin (H\&E) were captured using a virtual slide scanner with a maximum magnification of 40x, and 29 WSIs were used in the experiments. All images were taken from different patients and the sizes of the images were up to $54,000\times108,000$. 
To generate the training and test data, pathologists manually annotated the regions of five cancer subtypes: 1. Normal, 2. Lepidic, 3. Acinar/Papillary\footnote{Since it is quite difficult for even pathologists to identify Acinar and Papillary~\cite{YoshizawaA2011}, we put these two classes into one class.}, 4. Micropapillary, and 5. Solid, where \lq Normal\rq\, indicates the region of outside the tumors, and the other four classes are tumor subtypes. To segment the images into multi-subtype regions, they were first segmented using the two-class segmentation and the tumor regions were then segmented into subtype regions.
Figure~\ref{fig:Data} shows a typical WSI and the corresponding annotated mask image.
Some of the tumor regions cannot be identified with any subtype. These regions are shown in black, so in total there are six classes (four subtypes, normal and unclear labeled tumors).

\subsection{Training}
To train our AWMF-CNN model, we extracted a set of three different-magnification patches $\{X_i^1, X_i^2, X_i^3\}_{i=1}^{N}$, corresponding to the same regions, from the WSIs (Figure~\ref{fig:TradeOff}), where the window size was $256\times256$ pixels, the stride size was $256$ pixels, and the magnifications were (20x, 10x, 5x)\footnote{10x is the magnification with which pathologists usually check images for diagnosis.}. The corresponding scaled annotation mask images $\{T_{i,c}^{k}|k=1,...,3,c=1,...,M\}_{i=1}^{N_x}$ were used for the label data.
The image patches were randomly flipped along the horizontal axis and vertical axis for data augmentation.
We experimented using five-fold cross-validation; the 29 WSIs were divided into five sets. After that, each WSI image was split into image patches. The image patches of one set were used in the test, and the other patches were used for training.
We used 167,766 image patches for training the two-class segmentation and 20,848 image patches for training four-class segmentation.
The class ratios of the training images were [Normal, Tumor] = [0.67, 0.33] and [Lepidic, Acinar/Papillary, Micro Papillary, Solid] = [0.25, 0.29, 0.23, 0.23] at 20x magnification. 
Twenty percent of the randomly selected training data was used as validation data to prevent overfitting.
This validation data was also used as the training data $\bm{X}'$ for the weighting CNN.
In the weighting CNN, we used the second magnification images as the inputs in all experiments because this setup was slightly better than the case when all magnification images were used.
The five-fold data set was used to evaluate all the compared methods.
We used Nadam optimizer~\cite{dozat2016incorporating} with a learning rate of $10^{-4}$. The optimization was performed until 50 epochs or convergence. 

\subsection{Experimental results}
Figure~\ref{fig:Results} shows examples of the results in comparison. These were generated by overlaying multi-subtype segmentation results in the tumor region of the two-class segmentation.
The results of U-net, SegNet, Dilated-net, and Hard-Switch-CNN contain many fragmented regions (Figure~\ref{fig:Results} (c), (d), (e), and (g)).
Of the compared methods, DeepLabv3+ (f) produced best results.
Both of our methods gave better results than DeepLabv3+.
Their results in the two-class image segmentation task (Figure~\ref{fig:Results} (h) and (i)) were qualitatively similar, but the adaptive weight version was better than the fixed weight version in the semantic segmentation task, as shown in the enlarged images of Figure~\ref{fig:Results}.

We also evaluated three metrics; the overall pixel (OP) accuracy, the mean of the per-class (PC) accuracy, and the mean of the intersection over the union (mIoU) for two-class and four-class segmentation tasks. These metrics \cite{CsurkaG2013} are defined as: 
\begin{equation}
\operatorname{OP} = \frac{\sum_{c}{TP_c}}{\sum_c{(TP_c+FP_c)}},\hspace{3mm} \operatorname{PC} = \frac{1}{M}\sum_{c}{\frac{TP_c}{TP_c+FP_c}},\nonumber
\end{equation}
\vspace{-1.5mm}
\begin{equation}
\operatorname{mIoU} = \frac{1}{M}\sum_{c}{\frac{TP_c}{TP_c+FP_c+FN_c}},
\end{equation}
where $M$ is the number of classes, and $TP_c$, $FP_c$, and $FN_c$ are the numbers of true positives, false positives, and false negatives for class $c$, respectively. 

The performance metrics of each method are shown in Table~\ref{Table:accuracy2class} and~\ref{Table:accuracy4class}.
In the two-class segmentation task, both of our CNNs had better metrics in all case in comparison with U-net, SegNet, Dilated-net, and Hard-Switch-CNN.
Their performance and that of the best field of view (5x) of DeepLabv3+ were not significantly different.
As shown in the left image in Figure~\ref{fig:Venn_3experts_before}, each individual expert had high accuracy by itself, and the results of these experts were very similar.
In this case, we consider that the adaptive weighting strategy was not so effective, but the multi-field-of-view strategy improved the performance compared with those of the individual experts. Consequently, we consider that our methods had similar performance and were better than the others.

In the multiple-subtypes segmentation tasks, our AWMF-CNN achieved the best performance and the fixed version was second best. The improvement was larger than in the two-class segmentation task.
As shown in the right image in Figure~\ref{fig:Venn_3experts_before}, the results of these experts had the different regions and the union of the experts was much larger than the region identified by a single expert.
In this case, we consider that our AWMF-CNN can adaptively use the image features from the different-magnification images depending on the input image. This made it more accurate than the other methods.

\begin{figure*}[t]
\centering
\includegraphics[width=0.88\linewidth]{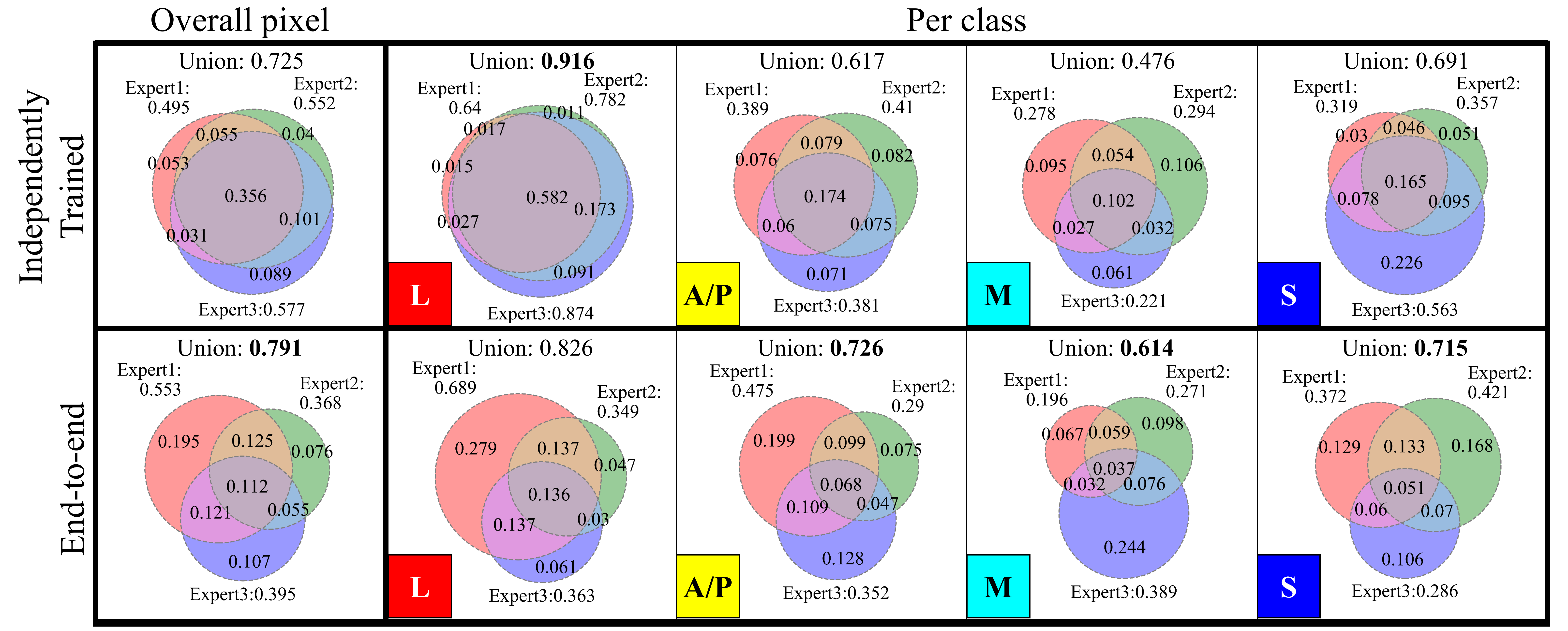}
\caption{Venn diagrams of the correct answer rates of individually pre-trained expert CNNs (Top) and after end-to-end learning (Bottom). Each column indicates the accuracy for each subtype; from left to right, average, Lepidic, Acinar/Papillary, Micropapillary, and Solid.}
\label{fig:venn_PerClass}
\vspace{-3mm}
\end{figure*}

\subsection{Changing expert CNNs in end-to-end training}
Figure~\ref{fig:venn_PerClass} shows the change in the correct answer rate for individually pre-trained expert CNNs (top) and after the end-to-end learning (bottom).
Through end-to-end learning, the union of the prediction results of the experts became large.
In particular, the end-to-end learning more improved the union of micro-papillary that had the smallest union than the others.
Although the circle of each expert became small on average after the end-to-end learning, it is considered that performance improved overall because all experts work complementarily.
Overall, the union of the correct answer rates was $7\%$ higher than that of the union of the individual experts.
We consider that this specialization of each expert by our model contributed to improving the performance.

\begin{table}[h]
\begin{center}
\caption{Comparison of two-class normal or tumor segmentations.}\label{Table:accuracy2class}
\vspace{-2mm}
\scalebox{0.85}[0.85]{
\begin{tabular}{ccccc}
\hline
Network & Magnification & OP & PC & mIoU \\ \hline\hline
U-net~\cite{U-net}\cite{RonnebegerO2015} & 20x & 0.890 & 0.876 & 0.774 \\
U-net~\cite{U-net}\cite{RonnebegerO2015} & 10x & 0.913 & 0.895 & 0.813 \\
U-net~\cite{U-net}\cite{RonnebegerO2015} & 5x & 0.910 & 0.899 & 0.810 \\
SegNet~\cite{BadrinarayananV2017}\cite{Segnet-implement} & 20x & 0.911 & 0.898 & 0.811 \\
SegNet~\cite{BadrinarayananV2017}\cite{Segnet-implement} & 10x & 0.909 & 0.902 & 0.810 \\
SegNet~\cite{BadrinarayananV2017}\cite{Segnet-implement} & 5x & 0.907 & 0.893 & 0.804 \\
Dilated-net~\cite{YuF2016} & 20x & 0.908 & 0.888 & 0.804 \\
Dilated-net~\cite{YuF2016} & 10x & 0.900 & 0.889 & 0.793 \\
Dilated-net~\cite{YuF2016} & 5x & 0.905 & 0.898 & 0.802 \\
DeepLabv3+~\cite{ChenLC2017_arXiv}\cite{DeepLabv3plus} & 20x & 0.911 & 0.894 & 0.811 \\
DeepLabv3+~\cite{ChenLC2017_arXiv}\cite{DeepLabv3plus} & 10x & 0.912 & 0.895 & 0.812 \\
DeepLabv3+~\cite{ChenLC2017_arXiv}\cite{DeepLabv3plus} & 5x & 0.917 & \textbf{0.915} & 0.825 \\
Hard-Switch-CNN~\cite{SamDB2017} & (20x,10x,5x) & 0.902 & 0.890 & 0.795 \\ \hline
Ours (Fixed) & (20x,10x,5x) & \textbf{0.921} & 0.907 & \textbf{0.831} \\
Ours (Adaptive) & (20x,10x,5x) & 0.916 & 0.904 & 0.821 \\ \hline
\end{tabular}
}
\end{center}
\vspace{-5mm}
\end{table}

\subsection{Varying expert networks}
To demonstrate that our AWMF-CNN can be adapted to any network, we trained it using U-net, SegNet, Dilated-net, and DeepLabv3+ as the expert networks for the subtype segmentation task; the training and test data sets were the same as in the above experiments.
Table~\ref{Table:another expert} shows that every AWMF-CNN trained by every type of expert was 3\% to 16\% more accurate than the corresponding individual network with the best magnification that produced the highest performance (Table~\ref{Table:accuracy4class}).

\begin{table}[h]
\begin{center}
\caption{Comparison of four-class subtypes segmentations.}\label{Table:accuracy4class}
\vspace{-2mm}
\scalebox{0.85}[0.85]{
\begin{tabular}{ccccc}
\hline
Network & Magnification & OP & PC & mIoU \\ \hline\hline
U-net~\cite{U-net}\cite{RonnebegerO2015} & 20x & 0.446 & 0.446 & 0.300 \\
U-net~\cite{U-net}\cite{RonnebegerO2015} & 10x & 0.484 & 0.481 & 0.331 \\
U-net~\cite{U-net}\cite{RonnebegerO2015} & 5x & 0.524 & 0.537 & 0.379 \\
SegNet~\cite{BadrinarayananV2017}\cite{Segnet-implement} & 20x & 0.477 & 0.477 & 0.320 \\
SegNet~\cite{BadrinarayananV2017}\cite{Segnet-implement} & 10x & 0.547 & 0.544 & 0.398 \\
SegNet~\cite{BadrinarayananV2017}\cite{Segnet-implement} & 5x & 0.492 & 0.525 & 0.326 \\
Dilated-net~\cite{YuF2016} & 20x & 0.433 & 0.422 & 0.274 \\
Dilated-net~\cite{YuF2016} & 10x & 0.445 & 0.456 & 0.314 \\
Dilated-net~\cite{YuF2016} & 5x & 0.515 & 0.528 & 0.378 \\
DeepLabv3+~\cite{ChenLC2017_arXiv}\cite{DeepLabv3plus} & 20x & 0.585 & 0.580 & 0.438 \\
DeepLabv3+~\cite{ChenLC2017_arXiv}\cite{DeepLabv3plus} & 10x & 0.625 & 0.624 & 0.474 \\
DeepLabv3+~\cite{ChenLC2017_arXiv}\cite{DeepLabv3plus} & 5x & 0.588 & 0.583 & 0.433 \\
Hard-Switch-CNN~\cite{SamDB2017} & (20x,10x,5x) & 0.486 & 0.484 & 0.347 \\ \hline
Ours (Fixed) & (20x,10x,5x) & 0.641 & 0.642 & 0.505 \\
Ours (Adaptive) & (20x,10x,5x) & \textbf{0.672} & \textbf{0.676} & \textbf{0.536} \\ \hline
\end{tabular}
}
\end{center}
\vspace{-5mm}
\end{table}

\begin{table}[h]
\begin{center}{
\caption{Segmentation accuracy of AWMF-CNN using other expert networks and improvement over the best individual expert.}
\label{Table:another expert}
\scalebox{0.85}[0.85]{
\begin{tabular}{clcc}
\hline
Expert network & Magnification & mIoU & improvement \\ \hline\hline
U-net~\cite{U-net}\cite{RonnebegerO2015} & (20x,10x,5x) & 0.536 & 0.157 \\
SegNet~\cite{BadrinarayananV2017}\cite{Segnet-implement} & (20x,10x,5x) & 0.459 & 0.061\\
Dilated-net~\cite{YuF2016} & (20x,10x,5x) & \textbf{0.537} & 0.159\\
DeepLabv3+~\cite{ChenLC2017_arXiv}\cite{DeepLabv3plus} & (20x,10x,5x) & 0.510 & 0.036\\ \hline
\end{tabular}
}
}\end{center}
\vspace{-6mm}
\end{table}

\section{Conclusion}
We proposed a novel Adaptive-Weighting-Multi-Field-of-View CNN that can adaptively use image features from different-magnification images depending on the input image to segment multiple cancer subtype regions in pathology. Our method mimics that pathologists check images by changing their magnification adaptively depending on the characteristic of the target image to make pathological diagnoses.
In experiments, we analyzed how each expert was specialized after the end-to-end learning; experts were re-trained to work complementarily, as a result, it increased the cases that either expert can predict a correct answer.
Our method outperformed state-of-the-art segmentation networks on multiple-subtypes segmentation tasks. 
We also showed that it can be applied to any type of network.
In addition to magnifications (experts), the importance of subtypes (channels) and locations (pixels) may depend on the characteristics of the input image.
In the future work, we will develop a method that can weight the combination of magnifications, subtypes, and locations.

\section*{Acknowledgments}
This work was partially supported by Technology and Innovation (Cabinet Office, Government of Japan) and JSPS KAKENHI Grant Number JP18H05104 and JP17K08740.

{\small
\bibliographystyle{ieee}
\bibliography{egbib}

\begin{thebibliography}{10}\itemsep=-1pt

\bibitem{AlsubaieN2018}
N.~Alsubaie, M.~Shaban, D.~Snead, A.~Khurram, and N.~Rajpootation.
\newblock A multi-resolution deep learning framework for lung adenocarcinoma
  growth pattern classific.
\newblock In {\em MIUA}, pages 3--11, 2018.

\bibitem{AltunbayD2010}
D.~Altunbay, C.~Cigir, C.~Sokmensuer, and C.~GunduzDemi.
\newblock Color graphs for automated cancer diagnosis and grading.
\newblock {\em IEEE Trans. Biomedical Engineering}, 57(3):665--674, 2010.

\bibitem{BadrinarayananV2017}
V.~Badrinarayanan, A.~Kendall, and R.~Cipolla.
\newblock Segnet: A deep convolutional encoder-decoder architecture for image
  segmentation.
\newblock {\em TPAMI}, 39(12):2481--2495, 2017.

\bibitem{Camelyon2017}
P.~Bandi, O.~Geessink, Q.~Manson, and M.~van Dijk{\it et.al.}
\newblock From detection of individual metastases to classification of lymph
  node status at the patient level: the camelyon17 challenge.
\newblock {\em TMI}, 2018.

\bibitem{Camelyon2016}
B.~E. Bejnordi, M.~Veta, P.~J. van Diest, and B.~van Ginneken~{\it et.al.}
\newblock Diagnostic assessment of deep learning algorithms for detection of
  lymph node metastases in women with breast cancer.
\newblock {\em JAMA}, 318(22):2199--2210, 2017.

\bibitem{ChangH2014}
H.~Chang, Y.~Zhou, A.~Borowsky, and K.~B. {\it et.al.}
\newblock Stacked predictive sparse decomposition for classification of
  histology sections.
\newblock {\em IJCV}, 113(1):3--18, 2014.

\bibitem{ChenH2016}
H.~Chen, X.~Qi, L.~Yu, and P.-A. Heng.
\newblock Dcan: deep contour-aware networks for accurate gland segmentation.
\newblock In {\em CVPR}, pages 2487--2496, 2016.

\bibitem{ChenLC2017_TPAMI}
L.~Chen, G.~Papandreou, I.~Kokkinos, K.~Murphy, and A.~Yuille.
\newblock Deeplab: Semantic image segmentation with deep convolutional nets,
  atrous convolution, and fully connected crfs.
\newblock {\em TPAMI}, 40(4):834--848, 2017.

\bibitem{ChenLC2017_arXiv}
L.~Chen, G.~Papandreou, and F.~Schroff.
\newblock Rethinking atrous convolution for semantic image segmentation.
\newblock In {\em arXiv}, 2017.

\bibitem{ChenLC2016_cvpr}
L.~Chen, Y.~Yang, J.~Wang, W.~Xu, and A.~Yuille.
\newblock Attention to scale: Scaleaware semantic image segmentation.
\newblock In {\em CVPR}, 2016.

\bibitem{CholletF2016}
F.~Chollet.
\newblock Xception: Deep learning with depthwise separable convolutions.
\newblock In {\em CVPR}, pages 1800--1807, 2017.

\bibitem{clevert2015fast}
D.-A. Clevert, T.~Unterthiner, and S.~Hochreiter.
\newblock Fast and accurate deep network learning by exponential linear units
  (elus).
\newblock In {\em ICLR}, 2015.

\bibitem{CruzRoaA2014}
A.~Cruz-Roa, A.~Basavanhally, F.~Gonzalez, and H.~G. {\it et.al.}
\newblock Automatic detection of invasive ductal carcinoma in whole slide
  images with convolutional neural networks.
\newblock In {\em SPIE Medical Imaging}, 2014.

\bibitem{CsurkaG2013}
G.~Csurka, D.~Larlus, and F.~Perronnin.
\newblock What is a good evaluation measure for semantic segmentation?
\newblock In {\em CVPR}, 2013.

\bibitem{DingH2018}
H.~Ding, X.~Jiang, B.~Shuai, L.~Qun, and G.~Wang.
\newblock Context contrasted feature and gated multi-scale aggregation for
  scene segmentation.
\newblock In {\em CVPR}, 2018.

\bibitem{dozat2016incorporating}
T.~Dozat.
\newblock Incorporating nesterov momentum into adam.
\newblock In {\em ICLR Workshop}, 2016.

\bibitem{EigenD2015}
D.~Eigen and R.~Fergus.
\newblock Predicting depth, surface normals and semantic labels with a common
  multi-scale convolutional architecture.
\newblock In {\em ICCV}, 2015.

\bibitem{GouldS2009}
S.~Gould, R.~Fulton, and D.~Koller.
\newblock Decomposing a scene into geometric and semantically consistent
  regions.
\newblock In {\em ICCV}, 2009.

\bibitem{HouL2016}
L.~Hou, D.~Samaras, T.~M. Kurc, and Y.~{\it et.al.}. Gao.
\newblock Patch-based convolutional neural network for whole slide tissue image
  classification.
\newblock In {\em CVPR}, pages 2424--2433, 2016.

\bibitem{HuJ2018}
J.~Hu, L.~Shen, and G.~Sun.
\newblock Squeeze-and-excitation networks.
\newblock In {\em CVPR}, 2018.

\bibitem{ioffe2015batch}
S.~Ioffe and C.~Szegedy.
\newblock Batch normalization: Accelerating deep network training by reducing
  internal covariate shift.
\newblock In {\em arXiv preprint arXiv:1502.03167}, 2015.

\bibitem{JampaniV2016}
V.~Jampani, M.~Kiefel, and P.~Gehler.
\newblock Learning sparse high dimensional filters: Image filtering, dense crfs
  and bilateral neural networks.
\newblock In {\em CVPR}, 2016.

\bibitem{DeepLabv3plus}
bonlime/keras-deeplab-v3-plus: Keras implementation of deeplab v3+ with
  pretrained weights.
\newblock \url{https://github.com/bonlime/keras-deeplab-v3-plus}.
\newblock (Accessed on 11/16/2018).

\bibitem{Segnet-implement}
Segnet model implemented using keras framework.
\newblock \url{https://github.com/imlab-uiip/keras-segnet}.
\newblock (Accessed on 11/16/2018).

\bibitem{U-net}
Keras implementation of u-net.
\newblock \url{https://github.com/aymanshams07/Ultra/blob/master/unet.py}.
\newblock (Accessed on 11/16/2018).

\bibitem{KohliP2009}
P.~Kohli and {\it et al.}.~P.H.~Torr.
\newblock Robust higher order potentials for enforcing label consistency.
\newblock {\em IJCV}, 82:302--324, 2009.

\bibitem{KongB2017}
B.~Kong, X.~Wang, Z.~Li, Q.~Song, and S.~Zhang.
\newblock Cancer metastasis detection via spatially structured deep network.
\newblock In {\em IPMI}, pages 236--248, 2017.

\bibitem{KrizhenvskyA2012}
A.~Krizhevsky, Sutskever, Ilya, Hinton, and G.~E.
\newblock Imagenet classification with deep convolutional neural networks.
\newblock In {\em Advances in neural information processing systems}, pages
  1097--1105, 2012.

\bibitem{KumagaiS2017}
S.~Kumagai, K.~Hotta, and T.~Kurita.
\newblock Mixture of counting cnns: Adaptive integration of cnns specialized to
  specific appearance for crowd counting.
\newblock In {\em arXiv}, volume abs/1703.09393, 2017.

\bibitem{LiM2018}
M.~Li, W.~Zuo, S.~Gu, D.~Zhao, and D.~Zhang.
\newblock Learning convolutional networks for content-weighted image
  compression.
\newblock In {\em CVPR}, 2018.

\bibitem{LinG2016}
G.~Lin, C.~Shen, A.~Hengel, and I.~Reid.
\newblock Efficient piecewise training of deep structured models for semantic
  segmentation.
\newblock In {\em CVPR}, 2016.

\bibitem{LongJ2015}
J.~Long, Shelhamer, Evan, and T.~Darrell.
\newblock Fully convolutional networks for semantic segmentation.
\newblock In {\em CVPR}, pages 3431--3440, 2015.

\bibitem{MousaviHS2015}
H.~Mousavi, V.~Monga, G.~Rao, and A.~U. Rao.
\newblock Automated discrimination of lower and higher grade gliomas based on
  histopathological image analysis.
\newblock {\em J. Pathology Informatics}, 2015.

\bibitem{PinheiroP2014}
P.~Pinheiro and R.~Collobert.
\newblock Recurrent convolutional neural networks for scene labeling.
\newblock In {\em ICML}, pages I--82--I--90, 2014.

\bibitem{QinY2018}
Y.~Qin, K.~Kamnitsas, S.~Ancha, and J.~N. {\it et.al.}
\newblock Autofocus layer for semantic segmentation.
\newblock In {\em MICCAI}, 2018.

\bibitem{RonnebegerO2015}
O.~Ronneberger, P.~Fischer, and T.~Brox.
\newblock U-net: Convolutional networks for biomedical image segmentation.
\newblock In {\em MICCAI}, pages 234--241. Springer, 2015.

\bibitem{SamDB2017}
D.~B. Sam, S.~Surya, and R.~V. Babu.
\newblock Switching convolutional neural network for crowd counting.
\newblock In {\em CVPR}, 2017.

\bibitem{SirinukunwattanaK2018}
K.~Sirinukunwattana, N.~Alham, C.~Verrill, and J.~Rittscher.
\newblock Improving whole slide segmentation through visual context - a
  systematic study.
\newblock In {\em MICCAI}, pages 192--200, 2018.

\bibitem{VemulapalliR2016}
R.~Vemulapalli, O.~Tuzel, M.~Liu, and R.~Chellappa.
\newblock Gaussian conditional random field network for semantic segmentation.
\newblock In {\em CVPR}, 2016.

\bibitem{WangD2016}
D.~Wang, A.~Khosla, R.~Gargeya, H.~Irshad, and A.~Beck.
\newblock Deep learning for identifying metastatic breast cancer.
\newblock In {\em arXiv}, 2016.

\bibitem{XuY2015}
Y.~Xu, Z.~Jia, Y.~Ai, and F.~Z. {\it et.al.}
\newblock Deep convolutional activation features for large scale brain tumor
  histopathology image classification and segmentation.
\newblock In {\em ICASP}, 2015.

\bibitem{XuY2017}
Y.~Xu, Z.~Jia, L.-B. Wang, and Y.~{\it et.al.}. Ai.
\newblock Large scale tissue histopathology image classification, segmentation,
  and visualization via deep convolutional activation features.
\newblock {\em BMC bioinformatics}, 18(1):281, 2017.

\bibitem{XuY2016}
Y.~Xu, Y.~Li, M.~Liu, and Y.~{\it et.al.}. Wang.
\newblock Gland instance segmentation by deep multichannel side supervision.
\newblock In {\em MICCAI}, pages 496--504. Springer, 2016.

\bibitem{YaoJ2012}
J.~Yao, S.~Fidler, and R.~Urtasun.
\newblock Describing the scene as a whole: Joint object detection, scene
  classification and semantic segmentation.
\newblock In {\em CVPR}, 2012.

\bibitem{YoshizawaA2011}
A.~Yoshizawa, N.~Motoi, G.~J. Riely, and C.~{\it et.al.}. Sima.
\newblock Impact of proposed iaslc/ats/ers classification of lung
  adenocarcinoma: prognostic subgroups and implications for further revision of
  staging based on analysis of 514 stage i cases.
\newblock {\em Modern pathology}, 24(5):653, 2011.

\bibitem{YuF2016}
F.~Yu and V.~Koltun.
\newblock Multi-scale context aggregation by dilated convolutions.
\newblock In {\em ICLR}, 2016.

\bibitem{ZhaoH2017}
H.~Zhao, J.~Shi, X.~Qi, X.~Wang, and J.~Jia.
\newblock Pyramid scene parsing network.
\newblock In {\em CVPR}, 2017.

\bibitem{ZhengS2015}
S.~Zheng, S.~Jayasumana, B.~Romera-Paredes, and V.~V. {\it et.al.}
\newblock Conditional random fields as recurrent neural networks.
\newblock In {\em ICCV}, 2015.

\bibitem{ZhouY2014}
Y.~Zhou, H.~Chang, K.~Barner, P.~Spellman, and B.~Parvin.
\newblock Classification of histology sections via multispectral convolutional
  sparse coding.
\newblock In {\em CVPR Workshop}, pages 3081--3088, 2014.

\end{thebibliography}
}

\end{document}